\documentclass{article}
\usepackage{ijcai25}
\usepackage{bm}
\usepackage{kpfonts}
\usepackage{soul}
\usepackage{url}
\usepackage[hidelinks]{hyperref}
\usepackage[utf8]{inputenc}
\usepackage[small]{caption}
\usepackage{graphicx}
\usepackage{amsmath}
\usepackage{amsthm}
\usepackage{booktabs}
\usepackage{algorithm}
\usepackage{algorithmic}
\usepackage{color,xcolor}
\usepackage[switch]{lineno}

\title{BlackGoose Rimer: Harnessing RWKV-7 as a Simple yet Superior Replacement for Transformers in Large-Scale Time Series Modeling

}

\author{
Li weile,
Liu Xiao\\
\emails
alic2591709191@gmail.com \\
liu.xiao.in@gmail.com
}

\begin{document}

\maketitle

\begin{abstract}
Time series models face significant challenges in scaling to handle large and complex datasets, akin to the scaling achieved by large language models (LLMs). The unique characteristics of time series data and the computational demands of model scaling necessitate innovative approaches. While researchers have explored various architectures such as Transformers, LSTMs, and GRUs to address these challenges, we propose a novel solution using RWKV-7, which incorporates meta-learning into its state update mechanism. By integrating RWKV-7's time mix and channel mix components into the transformer-based time series model Timer \cite{liu2024timer}, we achieve a substantial performance improvement of approximately 1.13x to 43.3x and a 4.5x reduction in training time with 1/23 parameters, all while utilizing fewer parameters. Our code and model weights are publicly available for further research and development.
 \href{https://github.com/Alic-Li/BlackGoose_Rimer}{$https://github.com/Alic-Li/BlackGoose\_Rimer$}

\end{abstract}

\section{Introduction}

\begin{figure}
    \centering
    \includegraphics[width=1\linewidth]{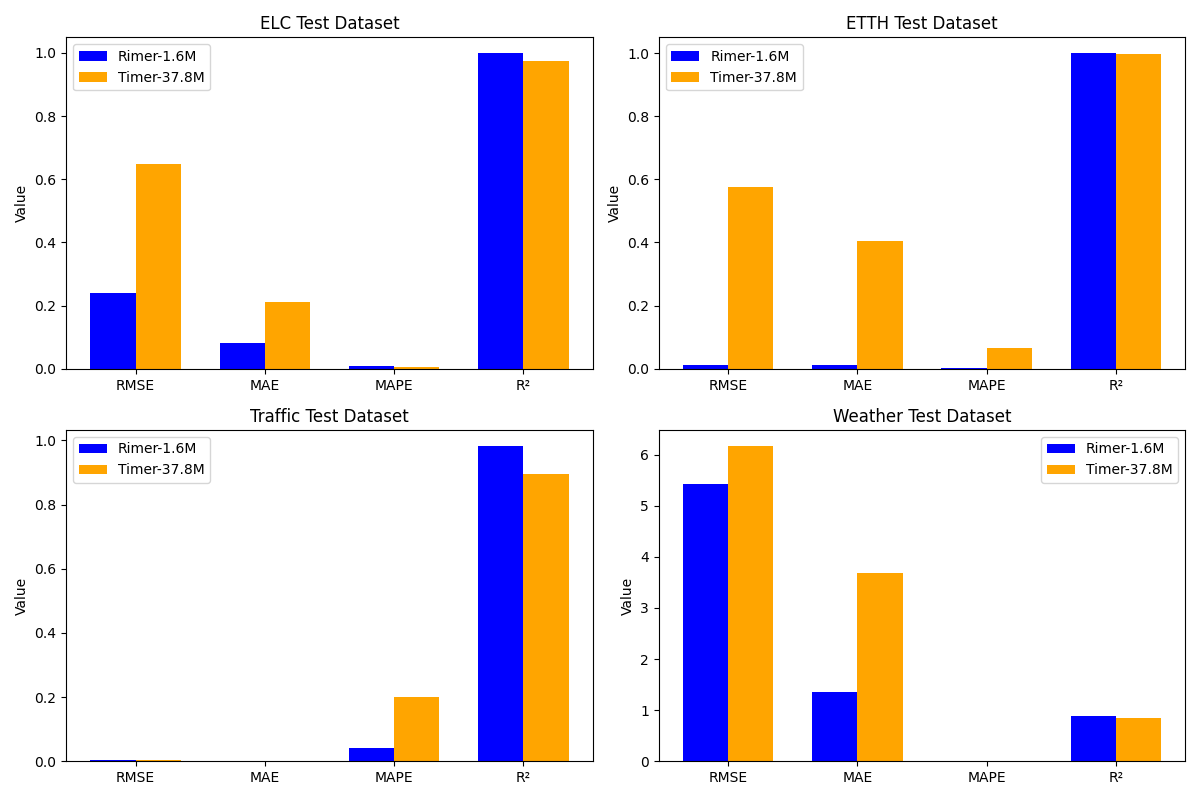}
    \caption{The benchmarks reveal that Rimer, with a significantly reduced parameter count of just 1.6 million, consistently outperforms or matches the performance of Timer, which relies on a much larger 37.8 million parameters, across multiple metrics. }
    \label{fig:enter-label}
\end{figure}

Time series modeling  stands as a fundamental pillar of machine learning \cite{wen2022transformers}, driving a wide array of applications such as forecasting, anomaly detection, and decision-making across fields like finance \cite{dong2024dft}, healthcare, and environmental science\footnote{This is an ongoing project}. With the rapid expansion of data in these domains, scaling time series models to efficiently process large datasets has emerged as a critical challenge. This issue parallels the scalability hurdles encountered in large language models (LLMs), where computational efficiency and performance at scale are paramount. However, the distinct properties of time series data—namely, temporal dependencies, high dimensionality, and often the demand for real-time analysis—pose unique difficulties that traditional approaches struggle to address effectively.
Over the years, researchers have investigated numerous architectures to tackle these challenges \cite{zhang2025tabulatime,lang5114041multi,wangchembr,hou2024rwkv,}, including transformers, Long Short-Term Memory networks (LSTMs), and Gated Recurrent Units (GRUs). While these models excel at capturing temporal patterns in smaller-scale settings, their efficiency and performance often degrade when applied to large datasets, highlighting the need for innovative solutions that can scale without compromising accuracy. This gap in capability has motivated the development of new methodologies tailored to the demands of large-scale time series modeling.
In this paper, we introduce a novel approach to meet these needs through RWKV-7, an architecture that integrates meta-learning into its state update mechanism \cite{grazzi2024unlocking}. RWKV-7 features two core innovations—time mix and channel mix \cite{peng2023rwkv} which enhance its ability to model complex temporal dynamics while maintaining computational efficiency. By embedding these components into the transformer-based time series model Timer, we achieve a remarkable 1.13x to 43.3x improvement in performance and a 4.5x reduction in training time compared to existing methods with 1/23 parameters. Furthermore, our approach leverages fewer parameters, offering a lightweight yet powerful solution for scaling time series models.

Our contributions in this work are twofold:

\begin{itemize}
\item Revisiting RNNs for Time Series with RWKV-7: We reintroduce the recurrent neural network (RNN) paradigm to time series modeling by adopting RWKV-7, a novel architecture tailored for large-scale time series data. This approach harnesses the sequential processing strengths of RNNs while overcoming their traditional challenges in handling extensive datasets, resulting in improved scalability and performance.

\item Benchmarking Against Transformer-Based Models: We perform comprehensive benchmarks to evaluate our RWKV-7-based model against previous transformer-based time series models. Our experiments highlight significant improvements, showcasing the effectiveness of our approach in terms of predictive accuracy and computational efficiency.

\end{itemize}


\section{Related Work}

\subsection{meta learning}
Meta-learning \cite{prudencio2004meta}, often described as "learning to learn," provides a powerful framework for enhancing large time series models by enabling them to quickly adapt to new tasks with limited data \cite{khosravi2023using}. These models, commonly used for tasks like forecasting and anomaly detection, typically require significant computational resources and extensive retraining when faced with new datasets. By incorporating meta-learning, they can utilize insights gained from prior tasks to minimize retraining efforts, employing techniques such as few-shot learning, transfer learning, and hyperparameter optimization to generalize effectively across varied scenarios. This approach not only boosts efficiency, versatility, and scalability but also tackles challenges like the need for diverse training tasks and initial computational overhead, paving the way for more adaptable and practical time series modeling solutions in real-world applications.

\subsection{Test time scaling}
In the era of big data, time series models have become essential tools across industries like finance and healthcare for tasks such as forecasting and anomaly detection. However, as the volume and velocity of time series data increase, efficiently scaling these models during the test phase—when predictions are made on new, unseen data—poses a significant challenge. Traditional methods often struggle with heightened computational demands and potential accuracy losses when processing longer sequences, higher-frequency data, or additional features. This paper presents a novel framework to address test time scaling \cite{sun2024learning} in time series models, utilizing adaptive computational techniques to maintain robust performance while preserving efficiency. Through extensive testing on diverse datasets, we showcase substantial gains in both speed and predictive accuracy, delivering a scalable solution for real-world applications. This work pushes forward the field of time series modeling and offers actionable insights for deploying these models in large-scale, real-time settings.

\section{Methodology}

\subsection{PRELIMINARIES}
\begin{figure}
    \centering
    \includegraphics[width=0.5\linewidth]{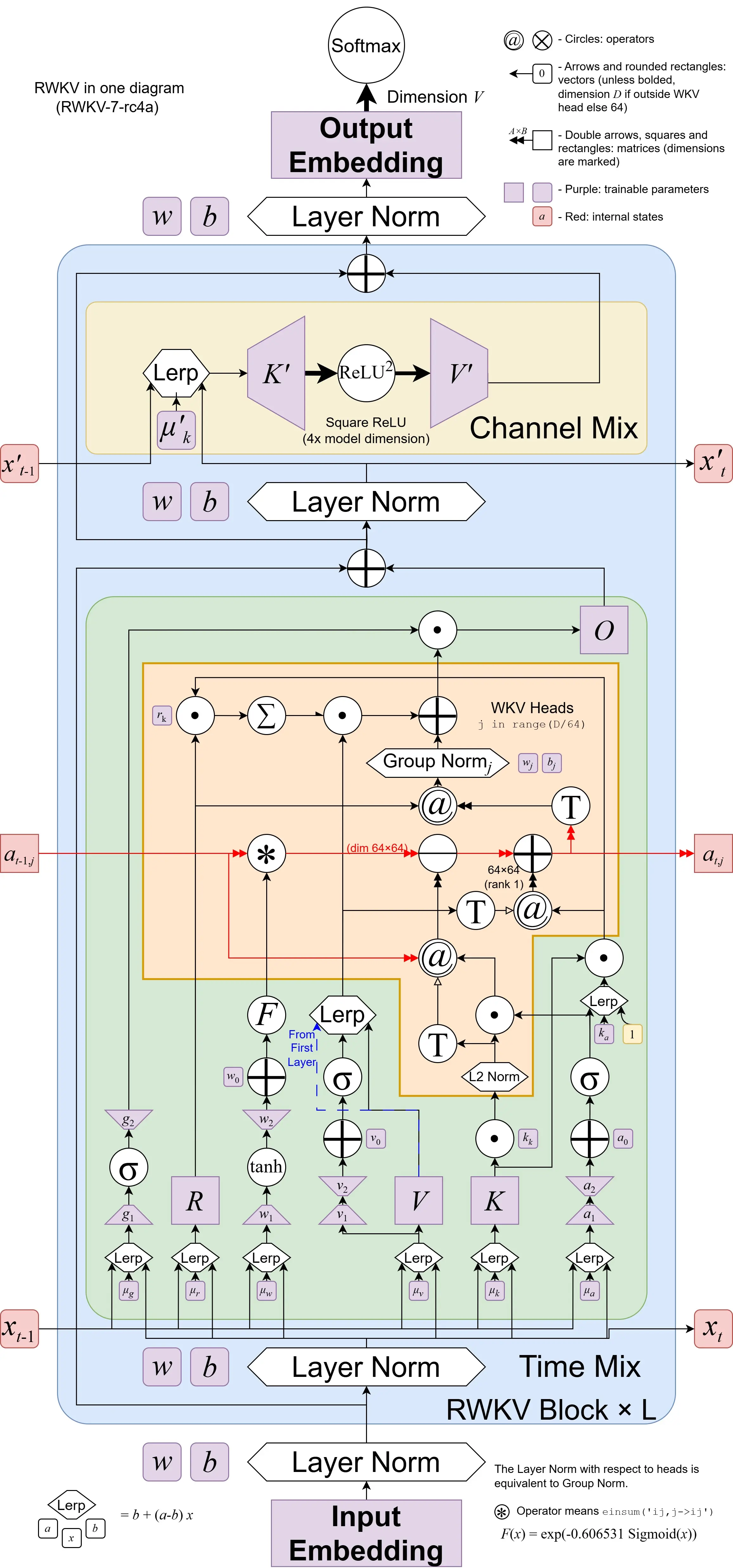}
    \caption{The RWKV-7 architecture is a RNN model that processes sequences using repeated RWKV blocks, each containing:1.A time mix block to blend current and past information.2.WKV heads for attention-like processing with an internal state to maintain memory.A channel mix block to transform the data further.}
    \label{fig:rwkv7}
\end{figure}

 The core structure of RWKV-7 (figure \ref{fig:rwkv7}) is dynamic evolution WKV state, have time mix and channel mix two main components.

the core state update rule:
\begin{equation}
    \textbf{State}_t = \textbf{State}_{t-1} \left({diag}(w_{t}) - \hat{\kappa}^T_t (a_t \cdot \hat{\kappa}_t)\right) + v^T_t \cdot \tilde{k}_t \cdot a_t 
\end{equation}

\subsection{Implicit RWKV-7 layers}
DEQ (Deep equilibrium models) \cite{bai2019deep} models aim to define layers as fixed-point solutions, where the output (or state) of a layer is the equilibrium point of a function, rather than being computed through a traditional sequential update. This is particularly useful for recurrent or transformer-like architectures, as it allows for "infinite-depth" behavior without explicitly unrolling the recurrence.

In a DEQ framework, our objective is to define $\text{State}_t$ (or a related hidden state $h_t$) as the solution to a fixed-point equation of the form:

\begin{equation}
    z * = f (z *, x)
\end{equation}

where $z^*$ is the equilibrium state, $x$ represents the input (including previous states and external inputs), and $f$ is a function (often a neural network layer) parameterized by learnable weights. For a recurrent setting, this can be adapted to incorporate temporal dynamics, where the fixed-point equation depends on both the current input and the previous state.

Instead of explicitly computing $\text{State}_t$ from $\text{State}_{t-1}$, we redefine the update implicitly. Let’s denote $h_t = \text{State}_t$, and consider the original equation as part of a transformation that we want to express as a fixed-point problem.
The original update can be written as:
\begin{equation}
h_t = f(h_t, \text{State}_{t-1}, x_t)
\end{equation}
 by introducing a self-recurrent term,
\begin{equation}
    h_t =  \phi\left(
    \begin{aligned}
    &W h_t + V \left( \text{State}_{t-1} \cdot (\text{diag}(w_t) - \kappa_t^T (a_t \cdot \hat{\kappa}_t)) \right)  \\
    & + U \left( v_t^T \cdot \kappa_t \cdot a_t \right) 
    \end{aligned}
    \right)
\end{equation}
where $W, V, U$ are Learnable weight matrices,Activation $\phi$ is relu to make non-linear.This DEQ form preserves the core dynamics state while reformulating it as an implicit layer,using latent-space iterations which can greatly improve expressive and efficiency.

\section{Evaluation}
To evaluate the performance of our proposed Rimer model against the baseline Timer, we conducted experiments across four diverse datasets,ELC, ETTH, Traffic, and Weather using metrics including RMSE, MAE, MAPE, and R², with datasets that are publicly available in our Rimer repository. Our Rimer model, featuring meta-learning in its state update rule and implemented with Triton operators, was trained on a Linux ROCm platform with AMD GPUs (Radeon Pro W7900 for training and Radeon RX6750XT for inference), achieving a 4.5x training time speedup compared to Timer due to its 1.6 million parameters versus the latter’s 37.8 million; this setup highlights our model’s unique strength in broad hardware compatibility, supporting training and inference across AMD GPUs, NVIDIA GPUs, and CPUs, a versatility enabled by Triton operators and developed on the ROCm platform.

\begin{table}
    \centering
    
    \begin{tabular}{lllll}
        \hline
        &RMSE   & MAE    & MAPE   & $R^2$  \\
        \hline
Timer-37.8M & 0.6488 & 0.2127 & 0.61\%  & 0.9755       \\
Rimer-1.6M        & 0.2409 & 0.0814 & 0.81\%  & 0.9991          \\     
        \hline
    \end{tabular}
    \caption{ECL}
    
\vspace{1cm}

    \begin{tabular}{lllll}
        \hline
        &RMSE   & MAE    & MAPE   & $R^2$  \\
        \hline
Timer-37.8M & 0.5770 & 0.4050 & 6.5\% & 0.9968     \\
Rimer-1.6M        & 0.0133 & 0.0112 & 0.16\%  & 0.9998          \\ 
        \hline
    \end{tabular}
    \caption{ETTH}
\vspace{1cm}

    \begin{tabular}{lllll}
        \hline
        &RMSE   & MAE    & MAPE   & $R^2$  \\
        \hline
Timer-37.8M & 0.0055 & 0.0015 & 19.94\% & 0.8955     \\
Rimer-1.6M        & 0.0025 & 0.0006 & 4.01\%  & 0.9838          \\ 
        \hline
    \end{tabular}
    \caption{Traffic}
 \vspace{1cm}   
    \begin{tabular}{lllll}
        \hline
        &RMSE   & MAE    & MAPE   & $R^2$  \\
        \hline
Timer-37.8M & 6.1765 & 3.6839 & 0.88\%  & 0.8411     \\
Rimer-1.6M        & 5.4311 & 1.3621 & 0.34\%  & 0.8794          \\ 
        \hline
    \end{tabular}
    \caption{Weather}
    
    \label{tab:plain}
\end{table}

The evaluation demonstrates that Rimer-1.6M , despite its significantly smaller size, delivers competitive or superior performance compared to the much larger Timer-37.8M model across multiple time series modeling tasks. This highlights its remarkable efficiency and effectiveness.

These results reveal that Rimer-1.6M, with just 1.6 million parameters, not only competes with but often outperforms Timer-37.8M, which relies on 37.8 million parameters. Its consistently higher $R²$ values across all datasets demonstrate a superior ability to capture temporal patterns, while its lower MAPE in ETTH and Traffic datasets suggests greater robustness to relative errors, potentially due to better handling of outliers or smaller values. The lightweight architecture of Rimer-1.6M over 23 times smaller than Timer-37.8M, also implies significant computational advantages, such as faster training and inference times, making it highly practical for large-scale applications.

In conclusion, the benchmark underscores Rimer-1.6M as a compelling alternative to traditional transformer-based models like Timer-37.8M. Its ability to achieve strong performance with a fraction of the parameters highlights the potential of RWKV-based architectures for efficient, effective, and scalable time series modeling, addressing both performance demands and resource constraints in real-world forecasting tasks.

\section{Conclusion}
The integration of RWKV-7 core components into the Timer architecture by replacing its transformer backbone has yielded a significant performance boost, as demonstrated through our experimental evaluations. This enhancement is attributed to the expressive and efficient nature of the RWKV-7 architecture, which, despite being fundamentally a type of recurrent neural network (RNN), effectively combines the strengths of traditional RNNs with the scalability and parallelization benefits typically associated with transformers. This hybrid design enables Rimer to outperform the original Timer across diverse datasets, leveraging its lightweight structure (1.6 million parameters versus 37.8 million) and broad hardware compatibility to achieve a 4.5x training time speedup. These results underscore the potential of RWKV-7 as a promising alternative for large-scale time series modeling, offering a compelling balance of performance, efficiency, and adaptability for future research and real-world applications.

\subsection{Future Work}
Looking ahead, we plan to focus our future efforts on several key directions to further advance the Rimer model. First, we aim to enhance its capability to handle long contexts by optimizing the state update mechanism to effectively capture extended temporal dependencies, potentially through advanced memory-augmented architectures. Second, we will explore the integration of latent space representations with state chains, allowing the model to learn compact, meaningful abstractions of the input sequence while maintaining state continuity across time steps, which could improve both efficiency and predictive accuracy. Finally, we intend to investigate hybrid models that combine RWKV-7 with other architectures \cite{yueyu2025arwkv}, such as transformers or convolutional neural networks, to leverage their complementary strengths for more robust time series forecasting. These explorations will pave the way for next-generation models capable of addressing complex, long-range dependencies and diverse application scenarios.

\appendix
\section*{Acknowledgments}
We extend our gratitude to Cao Qihuan for his generous guidance, as well as to Li Zhiyuan and the RWKV community for their contributions to RWKV-FLA.

\bibliographystyle{named}
\bibliography{ijcai25}

\end{document}